\documentclass[final]{cvpr}

\usepackage{times}
\usepackage{epsfig}
\usepackage{graphicx}
\usepackage{amsmath}
\usepackage{amssymb}

\usepackage{subfigure}
\usepackage{amsmath,bm}
\usepackage{booktabs}       
\usepackage{color}      

\usepackage{multirow}

\usepackage[ruled,vlined]{algorithm2e}
\usepackage[switch]{lineno}
\usepackage{comment}

\usepackage[pagebackref=true,breaklinks=true,colorlinks,bookmarks=false]{hyperref}

\def\cvprPaperID{1682} 
\def\confYear{CVPR 2021}

\begin{document}

\title{Towards Spatio-Temporal Video Scene Text Detection via Temporal Clustering}

\author{Yuanqiang Cai\footnotemark[1] ,
Chang Liu\footnotemark[1] ,
Weiqiang Wang\footnotemark[2], and Qixiang Ye \\\\
{University of Chinese Academy of Sciences, Beijing, China}\\
{\tt\small \{caiyuanqiang15, liuchang615\}@mails.ucas.ac.cn, \{wqwang, qxye\}@ucas.ac.cn }
}

\maketitle
\renewcommand{\thefootnote}{\fnsymbol{footnote}} 
\footnotetext[1]{Equal contribution} 
\footnotetext[2]{Corresponding author} 
\thispagestyle{empty}

\maketitle

\begin{abstract}
With only bounding-box annotations in the spatial domain, existing video scene text detection (VSTD) benchmarks lack temporal relation of text instances among video frames, which hinders the development of video text-related applications.
In this paper, we systematically introduce a new large-scale benchmark, named as \textit{STVText4}, a well-designed spatial-temporal detection metric (STDM), and a novel clustering-based baseline method, referred to as Temporal Clustering (TC).
\textit{STVText4} opens a challenging yet promising direction of VSTD, termed as ST-VSTD, which targets at simultaneously detecting video scene texts in both spatial and temporal domains.
\textit{STVText4} contains more than $1.4$ million text instances from $161,347$ video frames of 106 videos, where each instance is annotated with not only spatial bounding box and temporal range but also four intrinsic attributes, including \emph{legibility}, \emph{density}, \emph{scale}, and \emph{lifecycle}, to facilitate the community.
With continuous propagation of identical texts in the video sequence, TC can accurately output the spatial quadrilateral and temporal range of the texts, which sets a strong baseline for ST-VSTD.
Experiments demonstrate the efficacy of our method and the great academic and practical value of the \textit{STVText4}. The dataset and code will be available soon.
\end{abstract}

\section{Introduction}
\begin{figure}[t]
 \centering
 \includegraphics[width=1.0\linewidth]{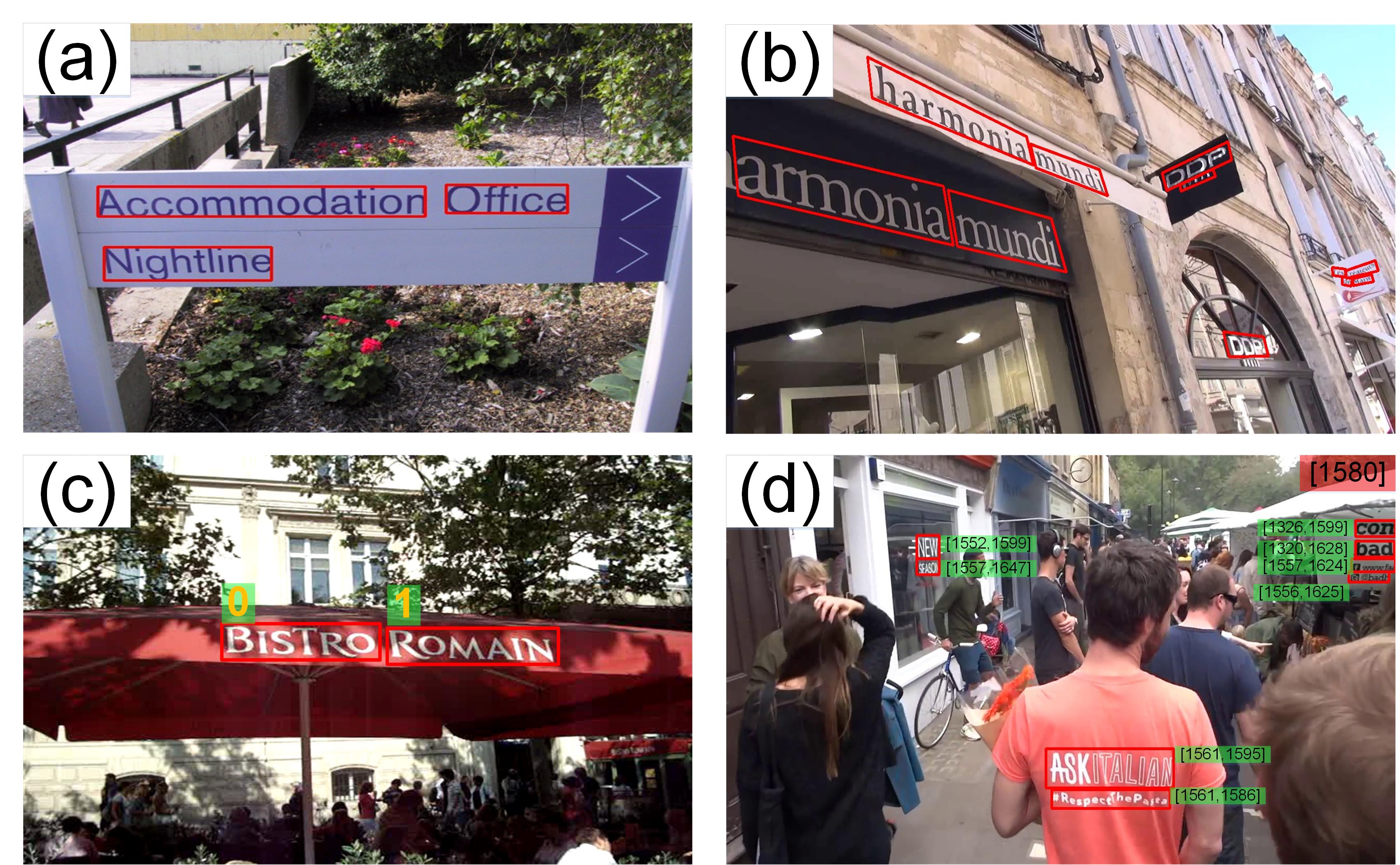}
 \caption{Comparison of annotations in typical benchmarks of (a) scene text detection in images~\cite{DBLP:conf/icdar/KaratzasSUIBMMMAH13}, (b) scene text detection in videos~\cite{DBLP:conf/icdar/KaratzasGNGBIMN15}, (c) scene text detection and tracking in videos~\cite{DBLP:conf/icip/MinettoTCLS11}, and (d) our \textit{STVText4} benchmark for the proposed ST-VSTD task. In (b), there are only spatial bounding-box annotations ignoring the temporal relation of identical text instances. In (c), the tracking id of texts are shown in left-top corner of each box. In (d), the right-top corner denotes the current time, and each text instance is annotated by spatial quadrilateral and temporal boundary [start time, end time]. Best viewed in color and zoom in.}
\label{fig:CompDiffTask}
\end{figure}

Video scene text detection (VSTD) is a fundamental task in computer vision, which aims to answer the question: ``where are the text instances in a video?''. It is extremely useful in the video-related applications, such as content-based video indexing, understanding, and summary in the tremendous video warehouses. However, similar to the box representation of text instances in image scene text detection benchmarks \cite{DBLP:conf/icdar/KaratzasSUIBMMMAH13, DBLP:conf/icdar/NayefYBCFKLPRCK17, DBLP:conf/iccv/WangBB11, DBLP:journals/corr/VeitMNMB16} (\ie, Fig. \ref{fig:CompDiffTask} (a)), existing VSTD benchmarks typically use only bounding boxes to represent text instances frame by frame (\ie, Fig. \ref{fig:CompDiffTask} (b)), ignoring the continuity and relation of the texts in the video sequence \cite{DBLP:conf/icdar/KaratzasGNGBIMN15, DBLP:conf/das/ZayeneSTHIA16, DBLP:journals/pami/TianYSH18}.
The video scene text detection and tracking benchmarks annotate the id of each text instance to build connections of identical text instances across the whole video. \cite{DBLP:conf/icip/MinettoTCLS11, DBLP:conf/mm/ChengLNP0Z19} (\ie, Fig. \ref{fig:CompDiffTask} (c)). Therefore, state-of-the-art scene text detectors typically struggle to constantly search and compare the text instances given in advance among adjacent video frames to localize them in the whole range of a video. As identical text instances from discontinual video segments with large temporal distance may have inconsistent semantic, multiple text instance tracking shows its limitations for the application scenarios.

\begin{table*}[t]
\footnotesize
\begin{center}
\setlength{\tabcolsep}{5.0pt}
\caption{Statistic comparison between our \emph{STVText4} dataset and other text detection datasets. If the dataset has a corresponding feature, it is marked as $\surd$, otherwise it is marked as $-$.}
\label{tab:StatisticsDiffData}
\begin{tabular}{c ccc ccc ccc cc cc}  
\toprule
\multirow{2}{*}{Dataset} &
\multicolumn{3}{c}{\#Image/frame} & \multicolumn{3}{c}{\#Text instance} &\multicolumn{3}{c}{\#Video} &\multicolumn{4}{c}{Attribute} \\
\cmidrule(lr){2-4} \cmidrule(lr){5-7} \cmidrule(lr){8-10} \cmidrule(lr){11-14}
 &train&test&all  &train&test&all &train &test &all  &Legibility&Density&Scale&Lifecycle\\
\midrule
ICDAR 2013 \cite{DBLP:conf/icdar/KaratzasSUIBMMMAH13} &229&233&462  &848&1,095&1,943  &-&-&- &-&- &- &-    \\
ICDAR 2015 \cite{DBLP:conf/icdar/KaratzasGNGBIMN15} &1,000&500&1,500  &11,886&5,230&17,116   &-&-&-  &$\surd$&- &-   &-    \\
MSRA-TD500 \cite{DBLP:conf/cvpr/YaoBLMT12} &300&200&500  &1,068&651&1,719    &-&-&- &$\surd$&- &-   &-     \\
COCO-Text \cite{DBLP:journals/corr/VeitMNMB16} &43,686&20,000&63,686  &118,309&27,550&145,859  &-&-&- &$\surd$&- &-  &-   \\
\midrule
AcTiV-D \cite{DBLP:conf/das/ZayeneSTHIA16} &1,480&363&1,843  &4,133&1,000&5,133   &4&4&$8$  &-&- &-   &-    \\
USTB-VidTEXT \cite{DBLP:journals/pami/TianYSH18} &9,839&17,831&27,670  &14,492&27,440&41,932  &1&4&$5$ &-&- &- &-    \\
ICDAR 2013 VT \cite{DBLP:conf/icdar/KaratzasSUIBMMMAH13} &9,790 &5,487 &15,277  &67,800 &26,134 &93,934  &13&$15$&28& $\surd$&- &-   &-   \\
LSVTD \cite{DBLP:conf/mm/ChengLNP0Z19} &37,160 &29,540 & 66,700 &340,805 &255,495 &596,300 & 66 & 34 &100&$\surd$&- &-   &-   \\
STVText4 (Ours) &\textbf{66,597}&\textbf{94,750}&\textbf{161,347} &\textbf{586,304}&\textbf{832,704}&\textbf{1,419,008} &59&$47$ &106 &$\surd$ &$\surd$  &$\surd$   &$\surd$     \\
\bottomrule
\end{tabular}
\end{center}
\end{table*}

In this paper, we develop the VSTD towards a challenging yet applicable setting, which is termed as spatial-temporal video scene text detection (ST-VSTD), targeting simultaneously detecting texts in both spatial and temporal domains. ST-VSTD emphasizes the partition of video scene based on the natural appearance and disappearance of texts in videos, that is, to predict the temporal boundary of each text instance. The temporal boundary of a text instance implies a relatively complete scene semantics. 
Specifically, we establish a large-scale ST-VSTD benchmark, referred to as \textit{STVText4}, which requires to detect the video scene text instances with both spatial bounding boxes and temporal ranges (\ie, start frame, and end frame). As shown in Fig. \ref{fig:CompDiffTask} (d), we link the identical text instances that appear continuously, and use minimum enclosing bounding boxes and a temporal interval, which consists of the start frame index and the end frame index, to indicate their spatio-temporal location. To the best of our knowledge, there has not been a benchmark for the proposed ST-VSTD task.

\textit{STVText4} contains more than $1.4$ million text instances from $161,347$ video frames of 106 videos.
For comprehensive and targeted evaluation and research, we not only expand the scale of the dataset (\ie, \#video, \#video frame, and \#text instance), and label the spatial quadrilateral location and temporal range of each text instance, but also additionally annotate four intrinsic attributes for each text instance, including \emph{legibility}, \emph{density}, \emph{scale}, and \emph{lifecycle}, Table~\ref{tab:StatisticsDiffData}.
Meanwhile, inspired by the commonly used evaluation protocols for conventional scene text detection and temporal action detection, we introduce a well-designed spatial-temporal detection metric (STDM). STDM penalizes algorithms in aspects of missing text instances, duplicate detection of identical instances, false positive detection, and false temporal location, offering fair, comprehensive, and reliable evaluation for ST-VSTD.

Due to various lifecycle length of different instances and huge representation differences of identical instances across video frames, it is challenging to predict the accurate temporal position of each text instance enclosed in bounding boxes. Inspired by the Density-Based Spatial Clustering of Applications with Noise algorithm (DBSCAN) \cite{ester1996density}, a novel clustering-based ST-VSTD approach, termed as temporal clustering (TC), is proposed to simultaneously detect video scene texts in the spatial and temporal domain. TC groups the same text instances in consecutive video frames to achieve the temporal text location. Extensive experiments demonstrate the effectiveness of our TC method for ST-VSTD on the proposed \textit{STVText4} benchmark. In summary, the contributions of this paper are four folds:

\begin{itemize}
\item A large-scale spatio-temporal video scene text detection (ST-VSTD) benchmark, referred to as \emph{STVText4}, with spatial bounding-box and temporal range annotations for each text instances, and additional four intrinsic attributes annotations including \emph{legibility}, \emph{density}, \emph{scale}, and \emph{lifecycle}, to facilitate the ST-VSTD community.

\item A well-designed spatio-temporal detection metric (STDM) which penalizes algorithms in aspects of missing text instances, duplicate detection of identical instances, false positive detection, and false temporal location, offering fair, comprehensive, and reliable evaluation for ST-VSTD.

\item A strong clustering-based baseline method for ST-VSTD, termed as temporal clustering (TC), which extends the text detector from the spatial domain to the temporal domain by using the continuous propagation of identical texts in the video sequence, to simultaneously output the spatial quadrilateral and the temporal range with the start and end boundaries.

\item Extensive experiments validate the effectiveness of the proposed method and the challenge aspects of the proposed benchmark, providing valuable insights for future research in the field of ST-VSTD.
\end{itemize}

\begin{figure*}[t]
\centering  
\includegraphics[width=6.9in,height=3.5in]{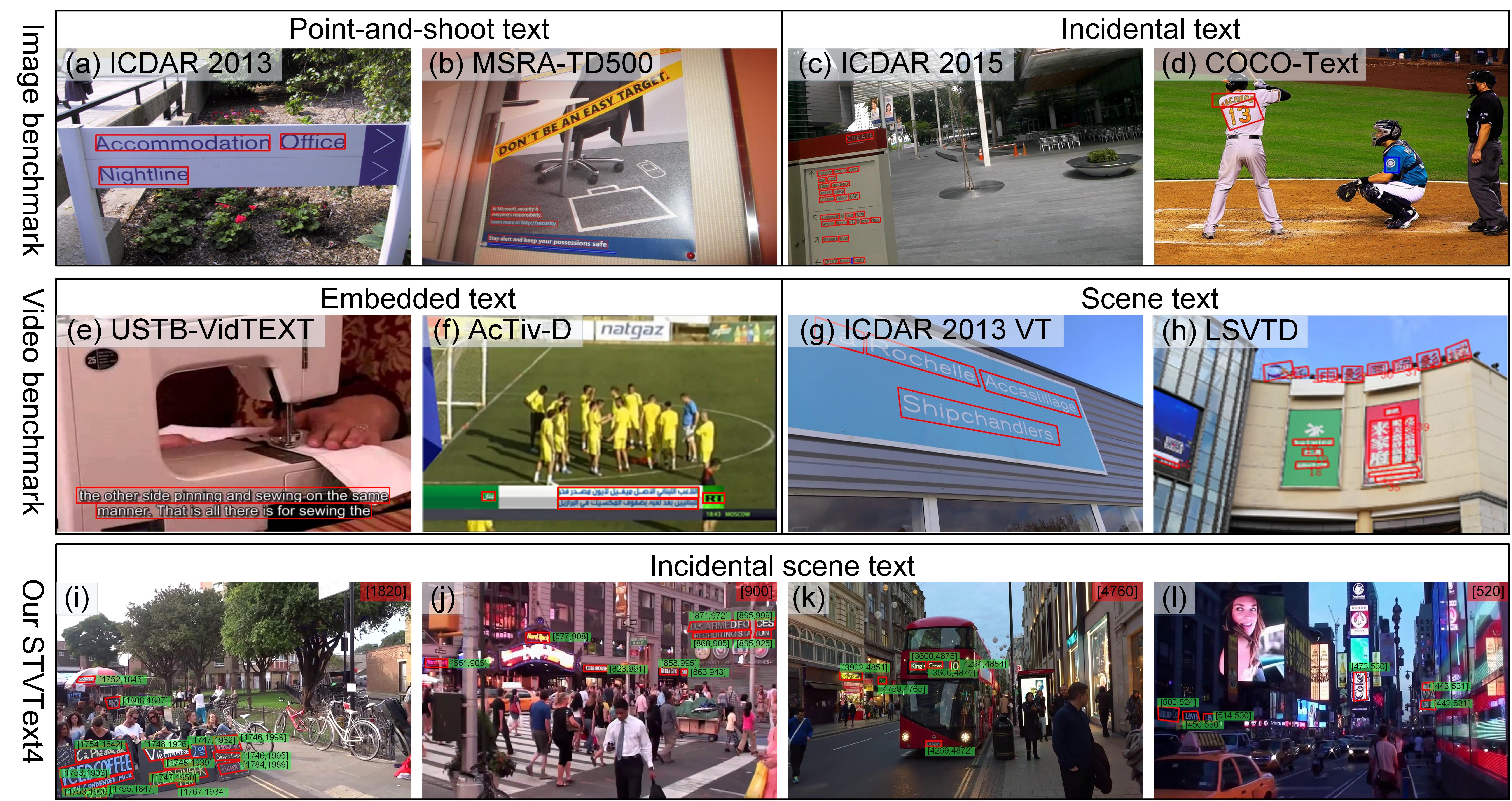}
\caption{ Existing image text benchmarks (\eg, (a)-(d) with point-and-shoot and incidental texts) and video text benchmarks (\eg, (e)-(h) with embedded and scene texts) focus on detecting texts in spatial domain. As shown in the last row ((i)-(l)), our \textit{STVText4} benchmark for the ST-VSTD task aims to detect video scene texts in the spatial and temporal domains simultaneously. Best viewed in color and zoom in.}
\label{fig:DataCompareShow}
\end{figure*}

\section{Related Works}

\subsection{Benchmarks}
Video text benchmarks can be basically classified into video embedded text benchmark \cite{DBLP:journals/pami/TianYSH18, DBLP:conf/das/ZayeneSTHIA16} and video scene text benchmark \cite{DBLP:conf/icip/MinettoTCLS11,DBLP:conf/icdar/KaratzasSUIBMMMAH13,DBLP:conf/icdar/KaratzasGNGBIMN15,DBLP:conf/mm/ChengLNP0Z19}.
Different from the image text benchmarks, such as ICDAR 2013 \cite{DBLP:conf/icdar/KaratzasSUIBMMMAH13} (Fig.\ \ref{fig:DataCompareShow} (a)), MSRA-TD500 \cite{DBLP:conf/cvpr/YaoBLMT12} (Fig.\ \ref{fig:DataCompareShow} (b)), ICDAR 2015 \cite{DBLP:conf/icdar/KaratzasGNGBIMN15} (Fig.\ \ref{fig:DataCompareShow} (c)), and COCO-Text \cite{DBLP:journals/corr/VeitMNMB16} (Fig.\ \ref{fig:DataCompareShow} (d)), with clear text, high resolution, and limited scenarios, video text benchmarks collected in the process of motion not only enrich the presentation of text instances and video scenarios, but also apply to the current era of streaming media.
Video embedded text benchmarks, such as USTB-VidTEXT \cite{DBLP:journals/pami/TianYSH18} (Fig.\ \ref{fig:DataCompareShow} (e)), and Activ-D \cite{DBLP:conf/das/ZayeneSTHIA16} (Fig.\ \ref{fig:DataCompareShow} (f)), etc, consist of text images where texts are embedded and overlaid in the frames of movies and TV series, which are usually utilized as a key cue for video skimming, web page detection, and detecting, skipping, and removing the commercial blocks.
Video scene text benchmarks, such as Minetto's dataset \cite{DBLP:conf/icip/MinettoTCLS11}, ICDAR 2013 Video Text (VT)~\cite{DBLP:conf/icdar/KaratzasSUIBMMMAH13} (Fig.\ \ref{fig:DataCompareShow} (g)), ICDAR 2015 Video Text (VT)~\cite{DBLP:conf/icdar/KaratzasGNGBIMN15}, and LSVTD~\cite{DBLP:conf/mm/ChengLNP0Z19} (Fig.\ \ref{fig:DataCompareShow} (h)), etc, contain scene texts in videos which are helpful for a variety of practical applications, such as driving assistance systems, assisting visually impaired people, etc.

However, these video benchmarks are only enriched in scale, scene, and text presentation, and still use the individual frame based evaluation strategy of image text detection, which can not meet the needs of large-scale industry-relevant video indexing, understanding, and summary applications. In this work, we establish a large-scale video scene text detection benchmark from the perspective of practical application which opens up a promising direction towards spatio-temporal video scene text detection.


\begin{figure*}[t]
  \centering
  \subfigure[]{
  \centering
	\includegraphics[width=0.23\textwidth]{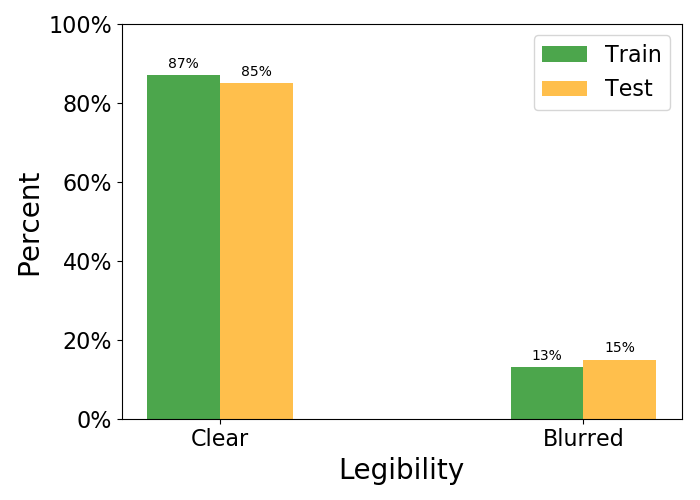}

  \label{fig:AS-legibility}
  }
  \subfigure[]{
  \centering
    \includegraphics[width=0.23\textwidth]{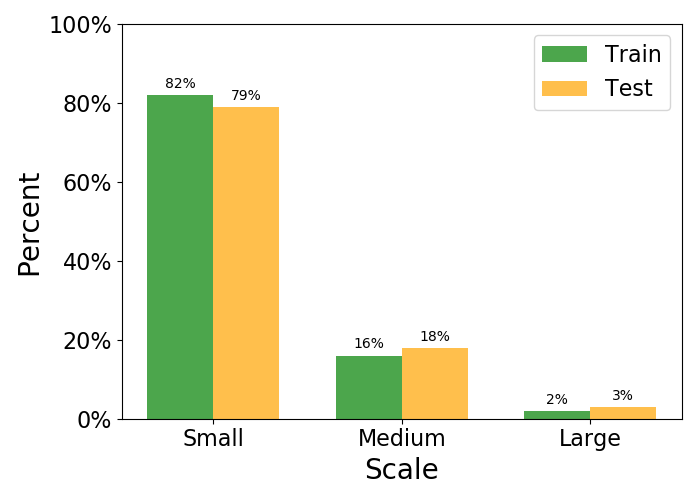}
  \label{fig:AS-scale}
  }
  \subfigure[]{
  \centering
    \includegraphics[width=0.23\textwidth]{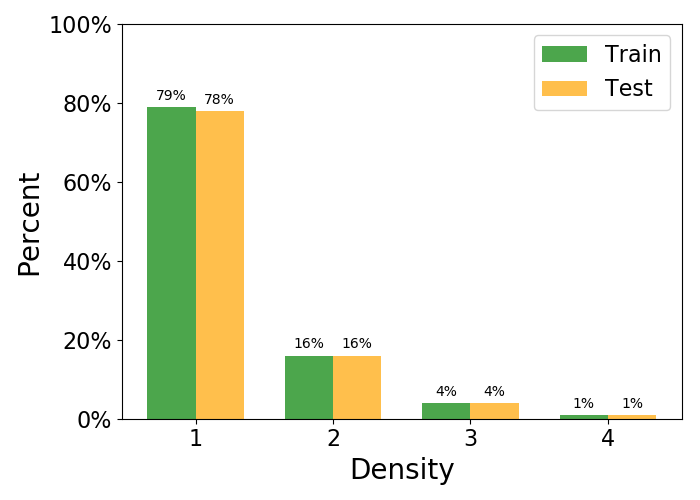}
  \label{fig:AS-density}
  }
  \subfigure[]{
  \centering
    \includegraphics[width=0.23\textwidth]{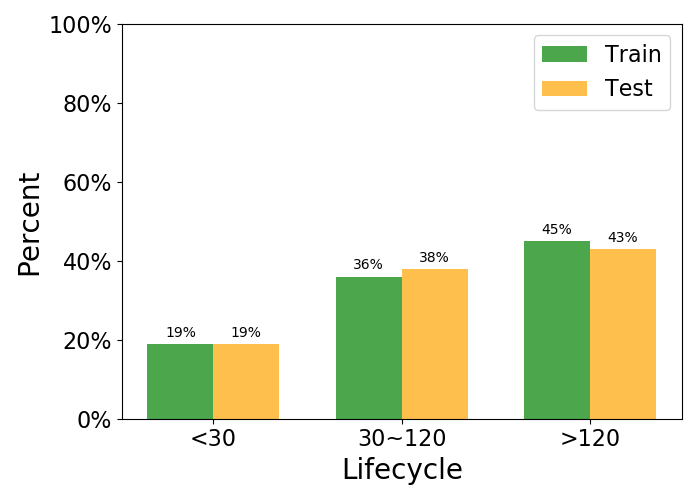}
  \label{fig:AS-lifecycle}
  }
\centering
  \caption{Illustration of attribute statistics of the {\em STVText4} dataset. (a) The text legibility distribution, (b) the text scale distribution, (c) the text density distribution, and (d) the text lifecycle distribution.}
\label{fig:Attribute-statistics}
\end{figure*}

\subsection{Methods}
Video text detection methods can be divided into three categories, \ie, single-frame enhancement, multi-frame fusion, text tracking.
\textit{Single-frame enhancement based video text detectors} firstly use the traditional description features (\eg, edge, color, and Laplace) to extract the text-like components, then classify these components by classifier and some rule constraints, finally combine the text components into a whole text \cite{DBLP:journals/pami/ShivakumaraPT11,
DBLP:journals/tip/LiangSLT15,DBLP:journals/pr/KhareSRB16}. Besides, after improving the single frames, text instances can be directly extracted by the text detectors \cite{DBLP:conf/aaai/LiaoWYCB20, DBLP:conf/cvpr/ZhangZHLYWY20, DBLP:journals/pr/CaiWCY20, DBLP:journals/tcsvt/YQCai20}.
\textit{Multi-frame fusion based video text detectors} focus on improving the quality of video texts and frames.
In \cite{DBLP:journals/tmm/LiuW12, DBLP:journals/pr/RoySJK0PL18}, authors firstly use multi-frame fusion to improve the quality of text instance and blur the background, then employ single-frame text detector to obtain the text location.
In \cite{DBLP:conf/mm/ChengLNP0Z19, DBLP:conf/mm/WangSWS19}, researchers firstly apply single frame processing technology to extract candidate texts, and then the redundancy information in temporal domain is used to improve the quality of video texts.
\textit{Text tracking based video text detectors} focus on enhancing the relation representation of text instances among adjacent frames. In \cite{DBLP:journals/tip/YangYPTZZY17, DBLP:conf/icdar/YuZLHDW19}, authors firstly use single-frame detection way to extract candidate text, then apply the persistence of text instance in temporal domain to track text instance with high confidence, and finally filter or merge the detected text instances and the tracking-propagative text instances in current frame.

The above video text detectors still stay in the spatial detection domain of a single or adjacent video frame, which does not overcome the representation difference of the same text in different video frames, nor can it locate the text in the temporal domain. Therefore, we propose a spatio-temporal video scene text detection method based on temporal clustering (TC), which can overcome the spatial difference of different video frames and solve the spatio-temporal localization problem by transforming the detection problem into a clustering problem.

\section{STVText4 Benchmark}

\subsection{Data collection and annotation}
For data collection, we invited $5$ experts to define the collection standards. Specifically, \emph{STVText4} is selected from the Internet video libraries and video datasets \cite{DBLP:conf/icdar/KaratzasSUIBMMMAH13} and \cite{DBLP:conf/icdar/KaratzasGNGBIMN15}. It contains more than $1.4$ million text instances from $161,347$ frames of 106 videos with resolution of $1280\times720$ pixels, Table~\ref{tab:StatisticsDiffData}. Also, the scenarios vary in a large range which span from day to night, e.g., Street view, Coffee corner, City road, Shopping mall, and City square. With unrestricted scenarios, various shooting modes, and uncertain imaging environments, Fig.\ \ref{fig:DataCompareShow}, our \emph{STVText4} spans several challenges, including multiple scales, multiple directions, dense text distribution, texture similarity, blurring, deformation, uneven illumination, and background clutter, etc. With the collected raw videos, we first chose the video clips with text, then make three rounds of screening to remove the ordinary videos. Then, the dataset is divided into training set with $586,304$ text instances in $66,597$ frames from $59$ videos, and testing set with $832,704$ text instances in $94,750$ frames from $47$ videos.

\begin{figure*}[t]
\centering
\includegraphics[width=1.0\linewidth]{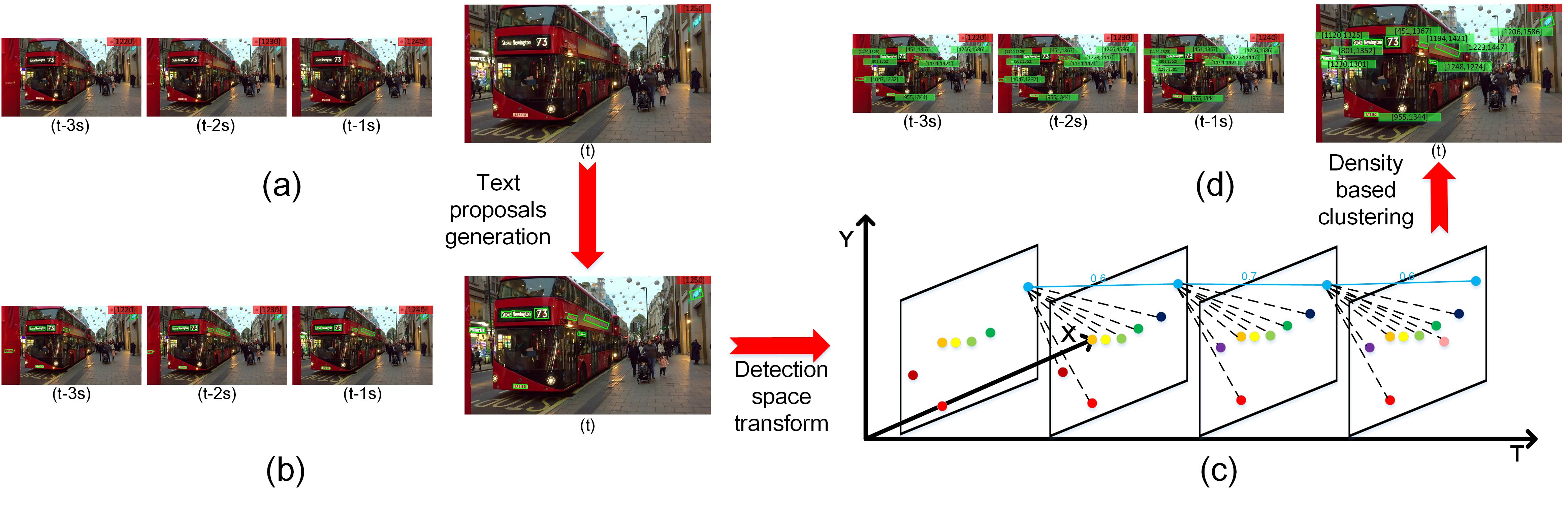}
\caption{The pipeline of the proposed spatio-temporal video scene text detection in videos. The right-top corner \emph{[current time]} denotes the current time of video frames in (a), (b), and (d). The green boxes indicate the spatial text location in (b) and (d). In (c), the solid line with value (\eg,$0.6$ and $0.7$) between two text points represents the IoU between two corresponding boxes, and the dotted line between two text points means that the two corresponding boxes do not intersect. The green mask with two values \emph{[start time, end time]} denotes the temporal boundary. Best viewed in color and zoom in.
}
\label{fig:pipeline}
\end{figure*}

For data annotation, we invited over 200 people to conduct three rounds of cross-checking for more than two months to ensure the annotation quality. Each text instance is labeled in the same quadrilateral way as in the ICDAR 2015 incidental text dataset \cite{DBLP:conf/icdar/KaratzasGNGBIMN15}. 
After the spatial location of the text instance is determined, the annotator will determine the temporal range of the text instance by browsing the length of the same text instance in the continuous video frame, and mark the label of start frame time and end frame time.
As we link the identical text instances among video frames to form the temporal range annotation, the proposed benchmark forces models to extract instance specific feature which make ST-VSTD a more academic and application valuable topic for the community. To our best knowledge, our \emph{STVText4} benchmark is the first one to emphasize the instance level temporal relation in the field of video scene text detection.

To provide the community with unified instance-level quantitative descriptions, and facilitate controlled evaluation for different approaches, we additionally define and annotate four attributes for each text instance, \ie, \emph{legibility}, \emph{density}, \emph{scale}, and \emph{lifecycle}.
Each attribute distribution of the {\em training} and {\em testing} subsets are illustrated in Fig.\ \ref{fig:Attribute-statistics}.

\textbf{Text legibility.} Due to the influence of motion, light, and distance during the video shooting, the legibility of different texts in the same video frame is different, even the same text in different video frames presents different legibility.
The detection results of the same text in adjacent video frames may be quite different, which can well verify whether the detector is robust to different sharpness texts.
To study the effect of text clarity on video scene text detection, we use subjective visual judgment to divide the text instances into two subsets, \ie, clear subset and blurred subset, that if the text instance is hard to be identified by the annotator, it belongs to the blurred subset, otherwise, the clear subset.

\textbf{Text scale.} Scale awareness is of indisputable importance for text detection. Text instances have quadrilateral annotation, and long bar presentation including horizontal distribution, vertical distribution, and random direction, which are different from the rectangular annotation for general object detection. Therefore, instead of the area of the text object, we divide the text scale into three subsets according to the length of the short side of each text instance, \ie, {\em small scale} subset ($<32$ pixels), {\em medium scale} subset ($32$-$64$ pixels), and {\em large scale} subset ($>64$ pixels).

\textbf{Text density.} Text instances often appear densely which usually makes text detectors confused to merge multiple text instances as one. Therefore, we grade text instances according to their density which can be used to finely uncover the influence of the text density to the detection performance. 
As the distance of adjacent text instances is typically sensitive to their scale and shape, we abandon distance based density measurements. Instead, we expand a text box that needs to be calculated by $0.1$ times of its short edge. Then we group the text instances according to the principle of whether the extended text box intersects with other text boxes. The text instance number in this cluster is the density of the expanded text instance.

\textbf{Text lifecycle.} The time length of a text instance appearing continually contains rich temporal information, which typically reflects the semantic importance of an text instance. Text lifecycle can be calculated as the difference of start frame time and end frame time of the text instance. More importantly, due to this temporal characteristic, the texts of video frames can be used to retrieve, understand, and summarize the video clip with the same texts. 
According to the continually existing time, we can divide the text instances into three subsets, \ie, {\em short lifecycle} subset ($<30$ video frames), {\em normal lifecycle} subset ($32\sim120$ video frames), and {\em long lifecycle} subset ($>120$ video frames).

\subsection{Evaluation protocol}
We design a new spatio-temporal detection metric (STDM) for ST-VSTD, which penalizes algorithms in the aspects of missing text instances, duplicate detection of identical instances, false positive detection, and false temporal location of detection. Inspired by the conventional scene text detection~\cite{DBLP:conf/icdar/KaratzasGNGBIMN15} and temporal action detection protocols~\cite{DBLP:conf/cvpr/HeilbronEGN15}, we evaluate detectors using the new designed metrics STDM, which consists of Precision (\ie, P$^{\it st}$), Recall (\ie, R$^{\it st}$), and F-score (\ie, F$^{\it st}$). Specifically, a correct spatio-temporal detection should satisfied two criteria, (1) the spatial detection intersection over union, $\text{IoU}^{\it s} = \frac{\widehat{B} \cap B^\ast}{ \widehat{B} \cup B^\ast}$, between the predicted bounding box $\widehat{B}$ and the ground-truth  $B^\ast$ is larger than the threshold $\theta_{\text{l}}$, \ie, $\text{IoU}^{\it s} \geq \theta_{\text{l}}$; and (2) the temporal localization intersection over union, $\text{IoU}^{\it t} = \frac{\widehat{R} \cap R^\ast}{ \widehat{R} \cup R^\ast}$, between the predicted time range $\widehat{R}$ and the ground-truth time range $R^\ast$ is larger than the threshold $\theta_{\text{r}}$, \ie, $\text{IoU}^{\it t} \geq \theta_{\text{r}}$. After that, if the best spatio-temporal match for a text instance meets the above two conditions at the same time, the text instance will be used as a correct detection. Precision and recall are defined as: $P^{\it st} = \frac{1}{n}\sum_{i=1}^{n}\frac{|H|}{|E|}$ and $R^{\it st} = \frac{1}{n}\sum_{i=1}^{n}\frac{|H|}{|T|}$, where $H$, $E$, and $T$ are the sets of correct detection, estimated, and ground-truth text instances in each video of \emph{testing} set, respectively.
The harmonic measure $F^{\it st}$ is adopted to combine the precision and recall, \ie, $F^{\it st}=\frac{1}{\frac{\alpha}{P^{\it st}}+\frac{1-\alpha}{R^{\it st}}}$, where the parameter $\alpha$ is usually set as $0.5$ to give equal importance to precision and recall.

\section{Temporal Clustering}
We introduce a clustering-based ST-VSTD method, termed as temporal clustering (TC), to tackle the lifecycle diversity of different text instances and the huge representation difference among the identical text instance in different video frames. Treating text instances as samples in a three-dimensional spatial-temporal solution space, TC transforms the detection problem to a clustering problem, which consists of three components, \ie, text proposals generation, detection space transform, density-based clustering, Fig. \ref{fig:pipeline}.

\begin{algorithm}[t]
  \caption{Spatio-temporal video scene text detection with density based clustering}
  \label{alg::densityBsdClustering}
  \LinesNumbered
    \KwIn{$D$: a dataset containing $m$ video frames with $n$ text instances, $\epsilon$: the search radius, $\tau_{d}$: the neighborhood distance threshold, $\tau_{l}$: the lifecycle threshold, and $\tau_{c}$: the cluster confidence threshold.}
    \KwOut{$D^{*}$: the optimal $D$.}

    \For{each frame $f$ in $D$}
    {    \For{each point $p$ in $f$}
          {
              \If{the $\epsilon$-neighborhood of $p$ has cluster $C_{i}$ and $dis(p,C_{i})<\tau_{d}$}
              {
                 add $p$ to the nearest cluster\;
                 update the cluster center, and the temporal boundary\;
              }
              \Else
              {
                 create a new cluster $C_{j}$, and add $p$ to $C_{j}$\;
              }

          }
    }

    \For{each cluster $C_{i}$ in $C$}
    {
        \If{$C^{l}_{i} <\tau_{l}$ and $C^{c}_{i} < \tau_{c}$}
              {
                 mark each point $p$ in $C_{i}$ and $D$ as noise, and delete the cluster $C_{i}$\;
              }
    }
    \For{each cluster $C_{j}$ in $C$}
    {
        \If{$C^{l}_{j} > C^{e}_{j}$}
        {
            complete the missing detection by two adjacent texts;
        }
    }
    \For{each cluster $C_{k}$ in $C$}
    {
        \For{each point $p$ in $C_{k}$}
        {
            add the temporal label to $p$ by $C^{l}_{k}$\;
            add the updated $p$ to the new data set $D^{*}$;
        }
    }

    \textbf{return} $D^{*}$;

\end{algorithm}
\vspace{-2mm}
\subsection{Text proposals generation}
The proposed spatio-temporal detection method is universal, that is, it can expand most of the image text detection algorithms, so that these algorithms can solve the proposed ST-VSTD task. Therefore, we choose several state-of-the-art text detection algorithms as our candidate text generators. Each frame in the video needs to be detected by the text detection algorithm. As shown in Fig.\ \ref{fig:pipeline} (b), the outputs of each candidate text box include: location information, text confidence, and video sequence time.

\subsection{Detection space transform}
After the text proposal generation, the algorithm gives both spatial and temporal detection information to each candidate text. However, these candidates are still in the original video space. They can only represent the content of the current video frame, not the whole lifecycle of each text. Thus we establish a three-dimensional detection representation space, Fig.\ \ref{fig:pipeline} (c). The \emph{X,Y} space denotes the spatial position of the text instance and the T-axis indicates the temporal position of the text instance.
For each candidate text, we reduce the candidate text box to a point in a new three-dimensional detection space. The spatial location of the point represents the spatial location of the geometry-center point of the candidate text. The confidence and temporal location of the point are the same as those in the original video space. The value on the line of points between video frames indicates the similarity between candidate text in adjacent video frames, that is, the IoU value of two candidate text boxes. To simplify the continuous representation of IoU, we only keep the line between each text instance and the instance in the adjacent video frame with the highest IoU.

\subsection{Density based clustering}
We transform the spatio-temporal text detection problem to a clustering problem. As shown in Fig.\ \ref{fig:pipeline} (c), in the three-dimensional spatio-temporal space, the distribution shape of points in the same cluster is streamline, so we adopt the density based clustering. Specifically, we use $C$ to represent a cluster (\eg, $C_{i}$, $C_{j}$, and $C_{k}$) with the same text points. The cluster is searched forward over time series, then the newly added point is regarded as the center point of the point cluster. The distance $dis(p_ {i},p_ {j})$ between two text points $p_{i}$ and $p_{j}$ indicates the similarity between the two text boxes $b_{i}$ and $b_{j}$. The larger the IoU between the two boxes is, the closer the distance between the text points will be. So $dis(p_ {i},p_ {j})$ can be calculated by $1.0-IoU(b_ {i},b_ {j})$. The distance $dis(p_{i},C_{k})$ between a point $p_{i}$ and a cluster $C_{k}$ can be obtained via calculating the distance between two text points, \ie, $dis(p_{i},C_{k})=Dis(p_{i},p_{k})$, where $p_{k}$ is the center point of the cluster $C_{k}$. The cluster $C_{i}$ refers to the set of existing instances of the same text. It will be updated to add the same text instance as the video frame time goes by. The $C^{e}_{j}$ is the number of text point in the cluster $C_{j}$. The confidence $C^{c}_{i}$ of cluster $C_{i}$ is the mean value of all confidences in the cluster. The lifecycle $C^{l}_{i}$ of a cluster $C_{i}$ is the time from the earliest text point to the latest text point. To understand the spatio-temporal detection with density based clustering easily, we give the pseudo code in Algorithm~\ref{alg::densityBsdClustering}.

\begin{figure*}[t]
\centering
\includegraphics[width=0.98\linewidth]{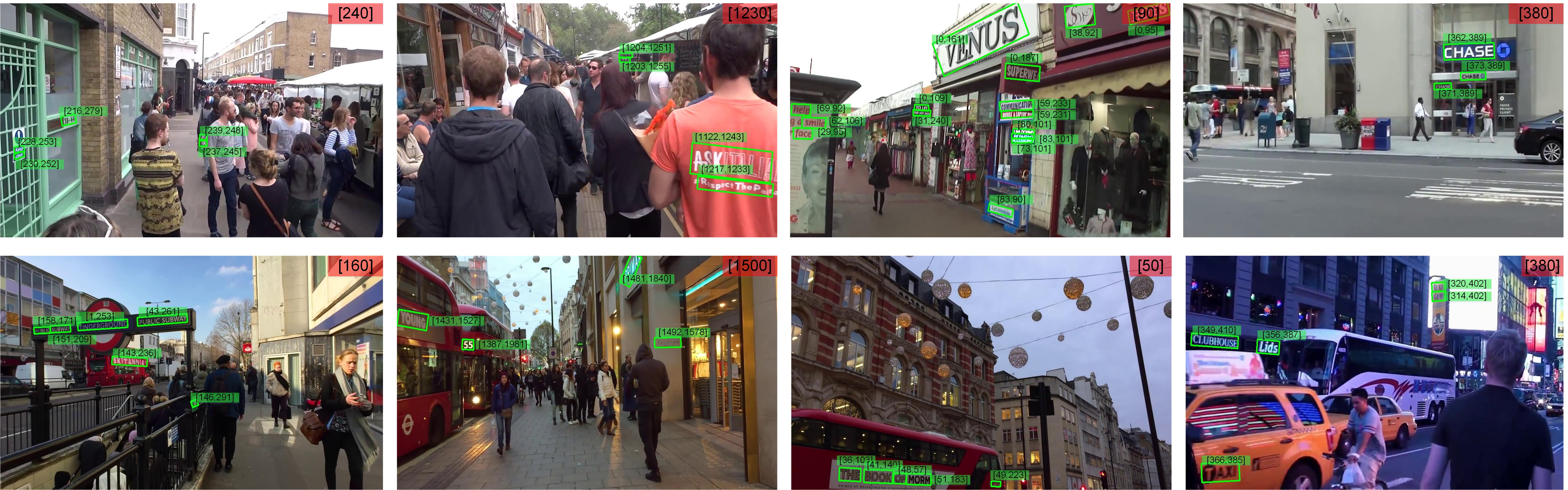}
\caption{Spatio-temporal detection results of TC(DB) on the proposed \emph{STVText4} benchmark. Each text instance has a spatiotemporal boundary to represent the scene information of the small fragments of videos. Best viewed in color and zoom in.}
\label{fig:Results}
\vspace{-2mm}
\end{figure*}

The proposed algorithm firstly group each text points into clusters to obtain the temporal boundary. To filter the false detection, we define and remove the noise text points by constraining the lifecycle and confidence of the cluster.
If the number of elements in the cluster is less than the lifecycle of the cluster, the text instances in the cluster is undetected in some video frames. To complete the missing detection $p_{i}$, we use the mean value of adjacent text instances before $p_{j}$ and after $p_{k}$ it to supplement, \ie, $p_{i}=Mean(p_{j},p_{k})$. Finally, as the temporal boundary of the cluster $C_{k}$ represents the temporal label of each text point $p$ in $C_{k}$, we can easily generate the updated text points with spatio-temporal location, Fig.\ \ref{fig:pipeline} (d).

\section{Experiments}

\subsection{Experimental Settings}
We train our method with state-of-the-art text detectors using the default VGG-16 and ResNet-50 backbone. The experiments are conducted on a single Titan Xp GPU and an Intel(R) Xeon(R) CPU E5-1603 v4 @ 2.80GHz. In evaluation protocol section, the spatial localization threshold $\theta_{l}$ is 0.5, the temporal range threshold $\theta_{r}$ is 0.5, and the balance parameter $\alpha$ is set as 0.5. In algorithm~\ref{alg::densityBsdClustering}, the search radius $\epsilon$ is 3, the neighborhood distance threshold $\tau_{d}$ is 0.7, the lifecycle threshold $\tau_{l}$ is 3, and the cluster confidence threshold $\tau_{c}$ is set as 0.3. All the parameters of state-of-the-art text detectors are adopt the default values.

\begin{table}[t]
\centering
\caption{Performance comparison on the proposed \emph{STVText4} benchmark, which is evaluated by the IC15 protocol \cite{DBLP:conf/icdar/KaratzasGNGBIMN15} and the proposed STDM protocol (denoted with superscript \emph{st}). Bold fonts indicate the best performance. Ditto for other tables.}
\label{tab:STDet-results}
\setlength{\tabcolsep}{3.0pt}
\begin{tabular}{l|ccc|ccc}
\hline
\multirow{2}{*}{Method}
&\multicolumn{3}{c|}{IC15 protocol}&\multicolumn{3}{c}{STDM protocol} \\
 \cline{2-7}
&P &R &F &P$^{\it st}$  &R$^{\it st}$ &F$^{\it st}$  \\
\hline
TC(R2CNN)  &\textbf{58.78}  &35.35 &44.15 &18.70 &23.32  &20.76\\
\hline
TC(IncpCF) &58.03  &37.96 &45.89 &18.02 &22.71  &20.10\\
\hline
TC(EAST) &57.93  &34.82 &43.49 &19.02 &21.96  &20.39\\
TC(PixelLink) &57.66 &37.60 &45.51 &17.01 &22.34 &19.31\\
TC(DB) &58.02  &\textbf{38.57} &\textbf{46.34} &\textbf{19.07} &\textbf{25.36}  &\textbf{21.77}\\
\hline
\end{tabular}
\vspace{-2mm}
\end{table}

\begin{table*}[t]
\centering
\caption{Evaluations on subsets of the proposed \textit{STVText4} divided by legibility (\emph{Clear} and \emph{Blurred}) and density ($1$, $2$, $3$, and $4$).}
\label{tab:Legibility-Density}
\setlength{\tabcolsep}{2.5pt}
\small   
\begin{tabular}{l|ccc|ccc|ccc|ccc|ccc|ccc}
\hline
\multirow{2}{*}{Method}
&\multicolumn{3}{c|}{Legibility (Clear)}&\multicolumn{3}{c|}{Legibility (Blurred)}
&\multicolumn{3}{c|}{Density (1)}&\multicolumn{3}{c|}{Density (2)}
&\multicolumn{3}{c|}{Density (3)}&\multicolumn{3}{c}{Density (4)}
\\
 \cline{2-19}
&P$^{\it st}$  &R$^{\it st}$ &F$^{\it st}$
&P$^{\it st}$  &R$^{\it st}$ &F$^{\it st}$

&P$^{\it st}$  &R$^{\it st}$ &F$^{\it st}$
&P$^{\it st}$  &R$^{\it st}$ &F$^{\it st}$
&P$^{\it st}$  &R$^{\it st}$ &F$^{\it st}$
&P$^{\it st}$  &R$^{\it st}$ &F$^{\it st}$\\
\hline

TC(R2CNN)
&34.74  &28.10 &31.07 &6.29 &5.99  &6.14
&\textbf{32.95}  &23.79 &27.63 &15.32 &20.70  &17.61
&6.11  &16.39 &8.91 &1.66 &14.23  &2.97\\
\hline
TC(IncpCF)
&33.90 &29.00  &31.26 &5.97  &6.28 &6.12
&31.43  &24.05 &27.25 &\textbf{16.58} &22.31  &\textbf{19.02}
&6.60  &19.74 &9.89 &1.86 &17.78  &3.37\\
\hline
TC(EAST)
&34.01  &27.04 &30.13 &6.07 &6.47  &6.26
&31.97  &23.33 &26.98 &14.48 &19.53  &16.63
&5.87  &18.63 &8.93 &1.85 &15.21  &3.30\\
TC(PixelLink)
&34.36  &29.75 &31.88 &6.47 &6.77  &6.62
&32.14  &24.88 &28.04 &15.85 &21.74  &18.33
&6.44  &18.31 &9.53 &2.02 &15.59  &3.58\\
TC(DB)
&\textbf{34.88}  &\textbf{30.51} &\textbf{32.55} &\textbf{7.02} &\textbf{6.81}  &\textbf{6.91}
&32.55  &\textbf{25.21} &\textbf{28.42} &16.54 &\textbf{22.29}  &18.98
&\textbf{7.27}  &\textbf{21.68} &\textbf{10.89} &\textbf{2.34} &\textbf{18.88}  &\textbf{4.16}\\
\hline
\end{tabular}
\end{table*}

\begin{table*}[t]
\centering
\caption{Evaluations on subsets of our \textit{ STVText4} divided by scale (\emph{Small}, \emph{Medium}, and \emph{Large}) and lifecycle (\emph{Short}, \emph{Normal}, and \emph{Long}).}
\label{tab:Scale-Lifecycle}
\setlength{\tabcolsep}{2.5pt}
\small   
\begin{tabular}{l|ccc|ccc|ccc|ccc|ccc|ccc}
\hline
\multirow{2}{*}{Method}
&\multicolumn{3}{c|}{Scale (Small)}&\multicolumn{3}{c|}{Scale (Medium)}
&\multicolumn{3}{c|}{Scale (Large)}&\multicolumn{3}{c|}{Lifecycle (Short)}
&\multicolumn{3}{c|}{Lifecycle (Normal)}&\multicolumn{3}{c}{Lifecycle (Long)}
\\
 \cline{2-19}
&P$^{\it st}$  &R$^{\it st}$ &F$^{\it st}$
&P$^{\it st}$  &R$^{\it st}$ &F$^{\it st}$
&P$^{\it st}$  &R$^{\it st}$ &F$^{\it st}$

&P$^{\it st}$  &R$^{\it st}$ &F$^{\it st}$
&P$^{\it st}$  &R$^{\it st}$ &F$^{\it st}$
&P$^{\it st}$  &R$^{\it st}$ &F$^{\it st}$\\
\hline

TC(R2CNN)
&31.47  &20.98 &25.18 &16.69 &29.83  &21.40
&\textbf{5.57}  &\textbf{22.24} &\textbf{8.91} &13.66 &10.23  &11.70
&25.28  &28.46 &26.77 &18.70 &23.32  &20.76\\
\hline
TC(IncpCF)
&31.34  &22.52 &26.21 &16.93 &27.47  &20.95
&5.02  &16.95 &7.75 &14.56 &11.77  &13.02
&24.67  &28.11 &26.28 &18.02 &22.71  &20.10\\
\hline
TC(EAST)
&30.93  &20.34 &24.54 &16.29 &26.51  &20.19
&5.33  &19.57 &8.37 &13.19 &10.27  &11.55
&24.95  &28.49 &26.60 &17.82 &21.59  &19.52\\
TC(PixelLink)
&31.58  &22.50 &26.28 &16.73 &30.94  &21.72
&4.96  &20.58 &7.99 &14.13 &11.27  &12.54
&\textbf{25.50}  &\textbf{29.36} &\textbf{27.29} &17.01 &22.34  &19.32\\
TC(DB)
&\textbf{32.31}  &\textbf{23.52} &\textbf{27.22} &\textbf{17.53} &\textbf{30.95}  &\textbf{22.38}
&5.18  &20.39 &8.26 &\textbf{15.17} &\textbf{12.02}  &\textbf{13.41}
&25.00  &28.78 &26.76 &\textbf{19.03} &\textbf{25.32}  &\textbf{21.73}\\
\hline
\end{tabular}
\vspace{-2mm}
\end{table*}

\subsection{Performance Comparison}
 We combine the proposed TC method with the state-of-the-art text detectors of three categories, \ie, anchor-based method (R2CNN \cite{DBLP:journals/corr/JiangZWYLWFL17}), tracking-based method (IncpCF \cite{DBLP:conf/pcm/WangWS18}), and anchor-free methods (EAST \cite{DBLP:conf/cvpr/ZhouYWWZHL17}, PixelLink \cite{DBLP:conf/aaai/DengLLC18}, and DB \cite{DBLP:conf/aaai/LiaoWYCB20}), for both conventional video text detection and the proposed ST-VSTD tasks, Table \ref{tab:STDet-results}. As conventional detectors can only predict spatial location frame by frame, we extend each algorithm with TC for ST-VSTD, and output the spatial and temporal location of each text instance simultaneously. The spatio-temporal detection results of TC(DB) on \emph{STVText4} are illustrated in Fig.~\ref{fig:Results}.

For the conventional video text detection task, we use the evaluation protocol in ICDAR 2015 \cite{DBLP:conf/icdar/KaratzasGNGBIMN15} to present the detection performance with Precision (P), Recall (R), and F-score (F).
Concretely, in terms of recall (R) metric, the performance of the anchor-free and tracking-based methods (\eg, EAST, PixelLink, DB, IncpCF) is better than that of the anchor-base method (\eg, R2CNN), while in terms of precision (P) metric, the performance is the opposite. Notably, the precision performance of the four algorithms is much better than the recall performance of them. It shows that the representation features of the text instances in the established benchmark are not salient, and the scenarios of the text instance are complex and clutter.

For ST-VSTD, we use the proposed protocol to indicate the performance of each algorithm, shown in Table \ref{tab:STDet-results}. Due to the consideration of both spatial and temporal boundary, the performance of all algorithms in spatio-temporal text detection is lower than that in spatial text detection (\eg, TC(DB), $58.02\%$ vs. $19.07\%$, $38.57\%$ vs. $25.36\%$, and $46.34\%$ vs. $21.77\%$). It demonstrates that the difficulty of spatio-temporal video text detection is far greater than that of conventional video text detection.
More important, we find that the attenuation of the precision score of all algorithms is particularly serious, specifically, $58.78\%$ $\to$ $18.70\%$ of TC(R2CNN), $58.03\%$ $\to$ $18.02\%$ of TC(IncpCF), $57.93\%$ $\to$ $19.02\%$ of TC(EAST), $57.66\%$ $\to$ $17.01\%$ of TC(PixelLink), and $58.02\%$ $\to$ $19.07\%$ of TC(DB). It shows that most of the detected text instances that meet the evaluation requirements of spatial location do not meet the evaluation requirements of temporal boundary, and implies that the difficulty of spatio-temporal detection is much higher than that of spatial location.
Besides, the comparisons between precision and recall of spatio-temporal detection is similar to that of spatial detection. In summary, the low performance of the state-of-the-art algorithms in the spatio-temporal domain substantiates the challenging aspect of the proposed \emph{STVText4}.

\subsection{Attribute analysis}
\textbf{Text legibility}.
To study the effect of the clarity of text presentation on the performance of spatio-temporal video scene text detection, we evaluate the clear and blurred texts respectively. As shown in the second and third columns of Table~\ref{tab:Legibility-Density}, the performance of all algorithms on clear text instance is much better than that on blurred text instance, \eg, $31.07\%$ vs. $6.14\%$ of TC(R2CNN), $31.26\%$ vs. $6.12\%$ of TC(IncpCF), $30.13\%$ vs. $6.26\%$ of TC(EAST), $31.88\%$ vs. $6.62\%$ of TC(PixelLink), and $32.55\%$ vs. $6.91\%$ of TC(DB). It shows that blurred text is an important factor restricting the overall performance of ST-VSTD.

\textbf{Text density}.
As shown in the last four columns of Table ~\ref{tab:Legibility-Density}, with the increase of text density, the detection performance of text gradually decreases (\eg, TC(DB), $28.42\%$ $\to$ $18.98\%$ $\to$ $10.89\%$ $\to$ $4.16\%$). It implies that dense text instances may interfere with each other, resulting in poor detection performance.

\textbf{Text scale}.
Due to the large number of training samples for small and medium texts, Fig. \ref{fig:AS-scale}, their detection performance is higher than the overall detection performance, \eg, TC(DB), $27.22\%$ $vs$. $21.77$ and $22.38\%$ $vs$. $21.77$, Table \ref{tab:Scale-Lifecycle} and Table \ref{tab:STDet-results}.
By comparing the precision (P$^{st}$) and recall (R$^{st}$) of each scale category, we find that small texts has higher detection precision, while large text with higher detection recall.
This is because the larger texts have more obvious characteristics to be distinguished, while being challenging for regression.


\textbf{Text lifecycle}.
In Table~\ref{tab:Scale-Lifecycle}, the detection performance of all algorithms for text instances with normal and long lifecycle is significantly better than that for text instances with short lifecycles (\eg, TC(DB), $26.76\%$ vs. $13.41\%$, and $21.73\%$ vs. $13.41\%$), and the detection performance of long lifecycle texts is slightly worse than that of the normal lifecycle ones. It is because that text instances appearing for a long time is easier to be detected. While with too long lifecycle, the variance of the representations of an identical text instance across video frames will increase, which will degenerate the temporal clustering results. These facts inspire us to improve the ST-VSTD performance by taking account of the lifecycle distribution of text instances.

\section{Conclusion}
We establish a large-scale spatio-temporal video scene text detection (ST-VSTD) benchmark, termed as \emph{STVText4}, with spatial bounding-box and temporal range annotations for each text instance, and additional four intrinsic attribute annotations to facilitate the ST-VSTD community, opening up a promising direction for video scene text detection. To implement ST-VSTD, we introduce a well-designed spatio-temporal detection metric (STDM) for fair, comprehensive, and reliable evaluation, and a strong clustering-based baseline for ST-VSTD, termed as temporal clustering (TC), which simultaneously outputs the spatial quadrilateral and the temporal range with the start and end boundaries. Extensive experiments on the proposed \emph{STVText4} not only validate the effectiveness of the TC method but also the challenging aspects of the proposed benchmark, which cast new insight into the video scene text detection area.

{\small
\bibliographystyle{ieee_fullname}
\bibliography{reference}
}

\end{document}